\documentclass[letterpaper, 10 pt, conference]{ieeeconf}  

\IEEEoverridecommandlockouts                              

\usepackage{graphics}
\usepackage{amsmath}
\DeclareMathOperator*{\argmax}{argmax}

\usepackage{amssymb}
\usepackage{epsfig} 
\usepackage{booktabs}
\usepackage{float}
\usepackage{url}
\title{\LARGE \bf
Adaptive Semi-Supervised Intent Inferral to Control a Powered Hand Orthosis for Stroke
}

\author{Jingxi Xu$^{1}$, Cassie Meeker$^{2}$, Ava Chen$^{2}$, Lauren Winterbottom$^{3}$, Michaela Fraser$^{3}$, Sangwoo Park$^{2}$ \\ Lynne M. Weber$^{3}$, Mitchell Miya$^{2}$, Dawn Nilsen$^{3, 4}$, Joel Stein$^{3,4}$, and Matei Ciocarlie$^{2,4}$
\thanks{This work was supported in part by the National Institute of Neurological Disorders and Stroke under grant R01NS115652.}
\thanks{$^{1}$Department of Computer Science, Columbia University, New York, NY 10027, USA. {\tt\small jxu@cs.columbia.edu}}
\thanks{$^{2}$Department of Mechanical Engineering, Columbia University, New York, NY 10027, USA. {\tt\small \{cgm2144, ava.chen, sp3287, mkm2201,
 matei.ciocarlie\}@columbia.edu}}
\thanks{$^{3}$Department of Rehabilitation and Regenerative Medicine, Columbia University, New York, NY 10032, USA. {\tt\small \{lbw2136, mgf2124, lw2739, dmn12, js1165\}@cumc.columbia.edu}}
\thanks{$^{4}$Co-Principal Investigators}
}

\begin{document}

\maketitle
\thispagestyle{empty}
\pagestyle{empty}

\begin{abstract}
In order to provide therapy in a functional context, controls for wearable robotic orthoses need to be robust and intuitive. We have previously introduced an intuitive, user-driven, EMG-based method to operate a robotic hand orthosis, but the process of training a control that is robust to concept drift (changes in the input signal) places a substantial burden on the user. In this paper, we explore semi-supervised learning as a paradigm for controlling a powered hand orthosis for stroke subjects. To the best of our knowledge, this is the first use of semi-supervised learning for an orthotic application.
Specifically, we propose a disagreement-based semi-supervision algorithm for handling intrasession concept drift based on multimodal ipsilateral sensing. We evaluate the performance of our algorithm on data collected from five stroke subjects. Our results show that the proposed algorithm helps the device adapt to intrasession drift using unlabeled data and reduces the training burden placed on the user. We also validate the feasibility of our proposed algorithm with a functional task; in these experiments, two subjects successfully completed multiple instances of a pick-and-handover task.

\end{abstract}

\section{Introduction}

Wearable devices can provide therapy in smaller, more frequent
aliquots than traditional robotic rehabilitation therapies, since they
have the potential to be administered outside of a clinical setting.
Furthermore, wearable robotics has the potential to provide therapy
while performing actual Activities of Daily Living (ADLs), which could
in turn make therapy more effective~\cite{krakauer2006}. However, in
order to realize this vision, wearable devices must be equipped with
user-driven controls that are both robust and intuitive.

\begin{figure}[t]
    \centering
    \includegraphics[width=0.98\linewidth]{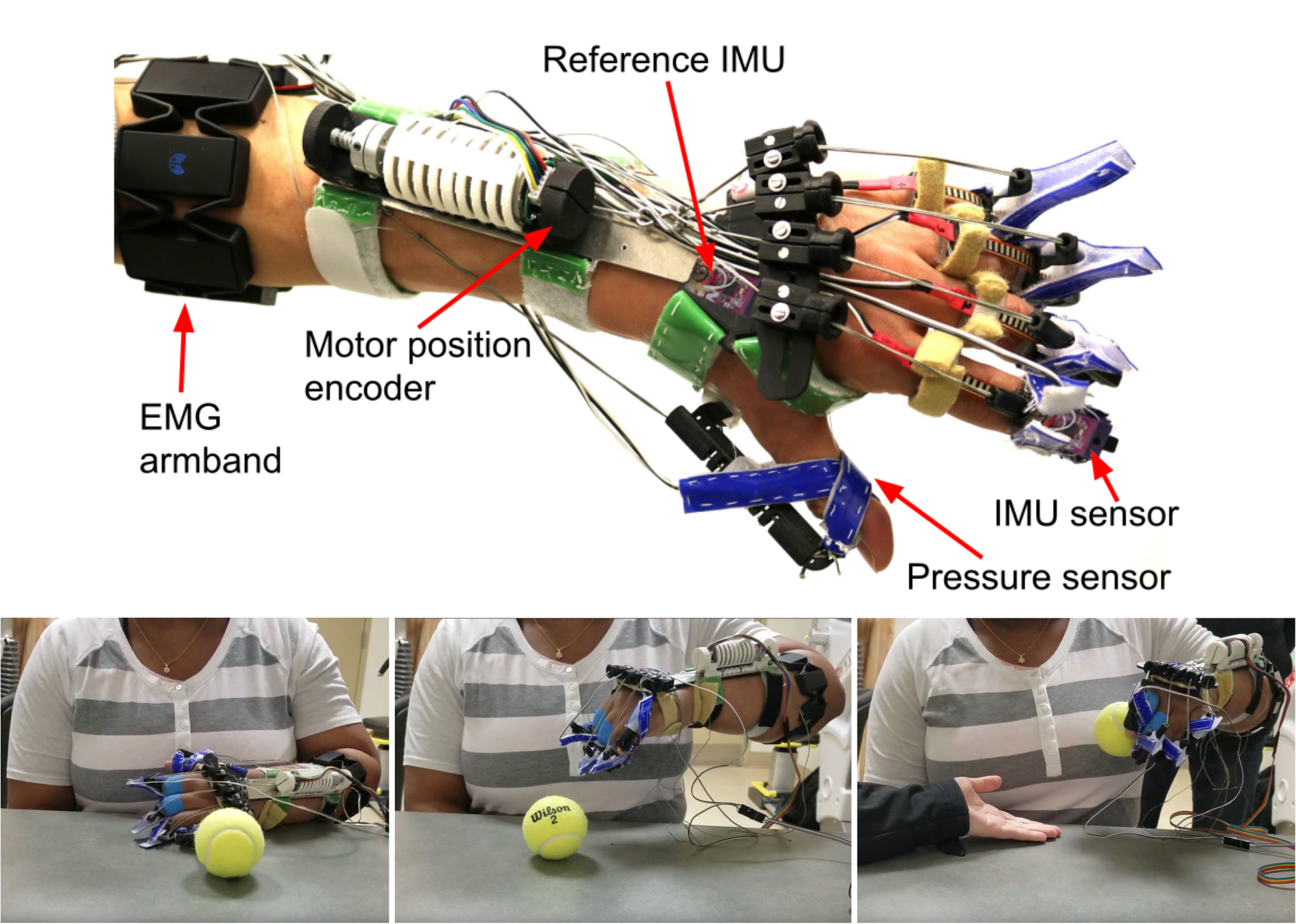}
    \caption{Top: hand orthosis with multimodal sensing suite. Bottom: stroke subject performing an assisted grasp. Due to abnormal synergies, muscle activation
    signals change significantly compared to collected training data, a type of concept drift that must be accounted for during intent inferral.}
    \label{fig:updated_orthosis}
\end{figure}

A common approach to user-driven control is to measure mechanical or
biological signals from the user in order to infer their intent. These
signals are commonly used in a Supervised Learning approach: a
regressor or a classifier is trained based on data with ground truth
intent labels associated with it. Such data is typically collected by
an experimenter in a dedicated training session. As an example, in our
own previous work, we have introduced a wearable hand orthosis
(Figure~\ref{fig:updated_orthosis}) for stroke subjects and used
ipsilateral electromyography (EMG) as an input to determine the
subject's intended hand motion~\cite{meeker2017}.

However, this approach is affected by challenges stemming from concept
drift, i.e. the phenomenon where the measured biosignals change over
time. Concept drift can occur between different sessions (intersession
drift) or gradually within a single session (intrasession
drift). In healthy subjects and amputees, concept drift is caused by
fatigue and repositioning of the sensors, among other
factors~\cite{jain2012improving}. In stroke subjects, concept drift is
additionally aggravated by abnormal muscle
coactivation~\cite{miller2012} and by the interaction of the hand and
the robotic device (Figure~\ref{fig:updated_orthosis}).

While we have found concept drift to be a significant problem for the stroke population, it is also rarely addressed in the literature. Supervised learners compensate for concept drift by using a training set that incorporates data with as much signal variation as possible. In our own previous work cited above, we trained with data for many arm poses and orthosis states.
However, such training comes at a high cost, since it is manually labeled, requiring the user to generate motions and labels specifically for training. This places a substantial burden on the user, especially in the case of stroke subjects, who fatigue quickly~\cite{riley2002changes}.



This paper focuses specifically on the intrasession drift with semi-supervised learning. Semi-supervised learning provides a potential solution, as it uses a small labeled dataset and then exploits additional unlabeled data to improve classifier performance. When intrasession drift happens, semi-supervised learners can adapt themselves to the changes in the input data. We hypothesize that these controls will require less training data than supervised controls, while maintaining high accuracies.

The main contributions of this paper are as follows. We propose a disagreement-based semi-supervised learning algorithm to help our orthosis adapt to intrasession concept drift when the device migrates, the subject gets fatigued or changes their arm poses, etc. To the best of our knowledge, \textit{we are the first to propose a semi-supervised control for a hand orthosis and validate its feasibility with functional tasks}. We evaluate the performance of our algorithm on data collected from five stroke subjects using multimodal sensing. We show that semi-supervised controls can adapt to intrasession drift with new unlabeled data and reduce the burden placed on the subject during training for ipsilateral hand controls. In the functional task experiments, two subjects successfully picked up and handed over a block multiple times in a minute with our proposed algorithm.



\section{Related Work}

In machine learning, there are three paradigms with which a classification algorithm can be trained: \textit{supervised learning}, where all of the training samples have a corresponding label, \textit{unsupervised learning}, where none of the training samples have a label, and \textit{semi-supervised learning}, where a small subset of the training samples have a corresponding label. 

Traditionally, controls for prosthetics and orthotics have been supervised --- trained on a relatively small dataset and used during a longer session~\cite{castellini2009, powell2013, lee2011}. However, leveraging unlabeled data may make the control algorithm more robust to fatigue, different arm poses, and abnormal muscle coactivation. 

Semi-supervised learning has been shown to be an effective control paradigm for hand prosthetics. In this field, semi-supervision assumes that the user will perform gestures at a low frequency to correct the classifier~\cite{jain2012improving, zhai2017self}. Semi-supervised formulations have been described for EMG classifiers, updating based on confidence, or when there are rapid changes in the prediction stream~\cite{sensinger2009adaptive}.
Other semi-supervised paradigms for EMG include using a post-processing neural network for prediction correction~\cite{amsuss2013self}, using all new data for updates~\cite{chen2013application}, reinforcement learning~\cite{edwards2016application}, and cycle substitution paired with probability weighting~\cite{zhang2013adaptation}. He et al. show that updating when confidence is high improves classification of wrist movement
~\cite{he2012adaptive}. Liu et al. find common characteristics among EMG classifiers  
to eliminate classifier retraining in later sessions~\cite{liu2015towards}.

All of the above works are specific to EMG. Many of them leverage the low-frequency nature of human hand motions in order to correct the classifiers~\cite{ jain2012improving, zhai2017self, sensinger2009adaptive}. The above works are designed for prosthetic controls and only tested on healthy subjects or amputees --- both populations have neuronormative muscle signaling. Only three studies look at intersession accuracy~\cite{zhai2017self, amsuss2013self, liu2015towards}.

To our knowledge, semi-supervised learning has never been applied to control for hand orthosis. Concept drift for stroke subjects differs from amputees because of the abnormal muscle coactivation in stroke subjects~\cite{miller2012}. Prosthetic controls do not have to contend with this phenomenon. 

There is a family of semi-supervision which we believe could also be applied to rehabilitation robotics. Disagreement-based semi-supervised learning asks multiple learners to collaborate in order to exploit unlabeled data~\cite{zhou2010semi}. Disagreement semi-supervision uses confident learners to train less confident learners and is well suited to ensemble learning~\cite{javed2005online, kolter2007dynamic, jiang2012semi, wang2006exploiting, xu2021active}. 

\section{Hardware}

Our orthosis is an exotendon device consisting of a forearm splint and fingertip components. These two parts are connected through an exotendon network. The device is underactuated, moving all of the fingers at once. When the motor retracts the tendons, the subject's fingers extend. When the motor releases the tendons, the subject uses their own grip strength to close their hand. 
Figure~\ref{fig:updated_orthosis} shows the exotendon device and the multimodal sensing suite, and additional details on the hardware can be found in our previous studies~\cite{park2018}. For the purposes of this work, we briefly review the sensing modalities mounted on our device:

\subsubsection{Forearm EMG} EMG signals of the forearm from an 8-sensor armband (Myo from Thalmic Labs).

\subsubsection{Motor Position}
A motor encoder provides position feedback, allowing us to determine the state of the orthosis.

\subsubsection{Finger Joint Angles}
Some subjects retain partial voluntary movement of their fingers, so we measure finger joint angles. We place one inertial measurement unit (IMU) on the pointer and one IMU on the back of the hand. The difference between the two IMUs measures the aggregate angle of the finger's joints. The derivative of the joint angle determines when the subject initiates voluntary motion or when the motor retracts or extends the exotendon network.

\subsubsection{Fingertip Pressure}
We measure the force between the device and the hand. This modality is provided by a force-sensitive resistor (FSR) mounted inside the thumb fingertip component. We use the time derivative of the pressure.

\section{Intent Detection}
\label{sec:intent_detection}
To determine the user's intent, we collect the raw EMG signals $(e^1 \dots e^8)$, motor position $\Gamma$, and time derivatives of joint $\Delta j$ and pressure $\Delta p$ data at 100Hz. At time $t$, we standardize the sensor data and collect it in a data vector $\boldsymbol x_t$:
\begin{gather}
\boldsymbol x_t = [e^1_t \dots e^8_t, \Gamma_t, \Delta j_t, \Delta p_t]^\top
\end{gather}

Our intent classifier is an ensemble with $\eta$ base learners:
\begin{gather}
CLAS^{ens}(\boldsymbol x_t) = \{CLAS^1(\boldsymbol f^1_t), ..., CLAS^\eta(\boldsymbol f^\eta_t)\} 
\end{gather}
where $\boldsymbol f^i_t$ represents the features for the base learner $i$. $\boldsymbol f^i_t$ may equal $\boldsymbol x_t$, or it may be a subset of features in $\boldsymbol x_t$. Each base learner $i$ outputs a vector of probabilities:
\begin{gather}
CLAS^i(\boldsymbol{f}^i_t) = \boldsymbol \Phi^i_t \\
\boldsymbol \Phi^i_t = [p_t^{R_i}, p_t^{O_i}, p_t^{C_i}]^\top
\end{gather} 
where $p_t^{R_i}$, $p_t^{O_i}$, and $p_t^{C_i}$ represent the probabilities that the user's intent is to relax, open and close their hand, respectively. Adding a relax intent allows the user to rest their arm occasionally and reduces the rate of fatigue. We note that $p_t^{R_i}, p_t^{O_i}, p_t^{C_i} \in [0, 1]$ and $p_t^{R_i} + p_t^{O_i} + p_t^{C_i} = 1$.

As such, $CLAS^{ens}(\boldsymbol x_t)$ will output a set of probabilities $\boldsymbol \Phi^{ens}_t$ which contains the probabilities predicted by each of the base learners in the ensemble:
\begin{gather}
\boldsymbol \Phi^{ens}_t = \{\boldsymbol \Phi^1_t, \boldsymbol \Phi^2_t, ..., \boldsymbol \Phi^\eta_t\} 
\end{gather}

We apply a median filter for each $\boldsymbol{\Phi}^i_t$ in $\boldsymbol{\Phi}^{ens}_t$, and then average the filtered $\boldsymbol{\hat{\Phi}}^i_T$ values:
\begin{gather}
\boldsymbol{\hat{\Phi}}^i_{T} = \texttt{MEDIAN}(\boldsymbol \Phi^i_t), t \in [T-0.25s, T] \\
\boldsymbol{\bar{\Phi}}^{ens}_{T} = \frac{1}{\eta} \times \sum_{i=1}^{\eta} \boldsymbol{\hat{\Phi}}^i_{T}
\end{gather}
We note that generally $\sum \boldsymbol {\hat{\Phi}}^i_{T} \neq 1$, and $\sum \boldsymbol{\bar{\Phi}}^{ens}_{T} \neq 1$.

We compare the maximum probability in $\boldsymbol{\bar{\Phi}}^{ens}_{T}$ to the corresponding threshold set by the experimenter $L^R$,  $L^O$, and $L^C$. If the maximum value does not exceed the threshold, the intent $I_T$ is set to the same prediction as the previous time step. Thresholds are manually adjusted according to user feedback such that control is responsive, but there are no spurious errors during sustained operation.

A motor command is then issued to the orthosis according to the predicted intent. If the classifier predicts that the user intends to open, the device retracts the tendon, extending the fingers. If the user intends to close, the device extends the tendon, allowing the user to flex the fingers. If the predicted intent is to relax, we continue to send the previous motor command to the device. 

\section{Semi-Supervised Learning for Intent Detection}

In this section, we discuss how to use a semi-supervised learning algorithm to improve our intent detection accuracies. Semi-supervised learning exploits unlabeled data to update the intent detection classifier. 
We hypothesize that this can make the intent detection robust to intrasession concept drift caused by fatigue, arm movement, device migration, etc., by constantly updating itself with new information. 

For any semi-supervised algorithm, we require an oracle. The oracle determines which new data sample should be used to update the classifier and labels the data. The data samples labeled by the oracle assemble into a training dataset. Since we use an ensemble classifier, we generate a training dataset $X^i$ for every base learner $i$ in the ensemble. When the aggregated dataset $X = \bigcup_{i=1}^{\eta}X_{i}$ contains a prespecified amount of data, we use the training data and labels from the oracle to update the classifier.

We would like our semi-supervised learning algorithm to address intrasession concept drift. Intrasession concept drift can be a sudden and large redistribution of the data in feature space (principally caused by subjects moving the arm to an unseen pose during training). It can also be a gradual shift in the data over time (primarily caused by device migration or subject fatigue). The proposed semi-supervised learning algorithm needs to be robust to these scenarios.

\subsection{Disagreement-based Oracle}
The semi-supervised learning algorithm presented here uses a disagreement-based oracle. When intrasession concept drift occurs, we hypothesize that the oracle can leverage the disagreement between multiple learners. Some of the base learners with a particular set of modalities will be more robust to changes in the data during the drift and will remain confident. Other learners which are not robust to the drift are corrected by the confident learners.

Our proposed disagreement-based semi-supervision is enacted as follows. At time $T$, we calculate the entropy $E^i_T$ of each base learner $i$:
\begin{gather}
E^i_T = - {\boldsymbol{\hat{\Phi}}^i_T}^\top log_k(\boldsymbol{\hat{\Phi}}^i_T)
\end{gather} 
where $k = 3$ is the number of possible intent classes. We use entropy as a measure of the learner's confidence. Lower entropy indicates higher confidence. 

Confident base learners are used to correct less confident learners. We define confident learners as those whose \mbox{$E^i_T < 0.2$.} Learners which are not confident have $E^i_T > 0.8$. We select our confidence thresholds empirically. 

If all confident learners agree on the subject's intent (i.e., if $\argmax(\boldsymbol{\hat{\Phi}}^i_T)$ is the same for all confident learners), for each unconfident learner $i$, we add $\boldsymbol{f}^i_T$ to $X^i$, along with the intent label agreed upon by the confident learners.

Once the combined number of training samples across all $X^i$s is a sufficiently large value (we choose 200 as the threshold), the data and labels are used to update the classifier, the supervision process starts, and all $X^i$s are reset.

Oracles can also correct the classifier prediction (before the base learners are updated). With disagreement-based semi-supervision, we only want confident learners to contribute to the final output of the ensemble. However, only including confident learners whose $E^i_T < 0.2$ is too restrictive, so we have an additional empirically-selected threshold for the correction. We calculate the final probability for the ensemble as an average of all the probabilities from learners whose entropy is less than $0.6$:
\begin{equation}
\boldsymbol{\bar{\Phi}}^{ens}_{T}=\frac{1}{\eta} \times \sum\limits_{\substack{i=1 \\ E^i_T < 0.6}}^{\eta} \boldsymbol{\hat{\Phi}}^i_{T}
\end{equation}
If there are no base learners whose $E^i_T < 0.6$, then $\boldsymbol{\bar{\Phi}}^{ens}_{T}$ is calculated using all base learners in the ensemble.
 
We are the first to use disagreement-based semi-supervision for an assistive robot. We can successfully leverage this paradigm for a novel application because our orthosis includes a multimodal sensing suite with independent sensing modalities~\cite{park2019}. Disagreement-based semi-supervision works best if the base learners include multiple independent views~\cite{blum1998combining}, or include a large number of base learners~\cite{zhou2005tri}. Therefore, we use ensembles with at least five base learners whose features are sampled randomly from all the sensors in the multimodal sensing suite.

\subsection{Updating the Classifier}

Once the aggregated dataset $X$ across all $X^i$s has a sufficient number of data samples, we update our classifier. We use linear discriminant analysis (LDA) for our base learners because it does not need past training data for updates. An LDA base learner $i$ with $k$ classes has the following parameters: a mean vector for each class $\boldsymbol \mu^i_k$ and a covariance matrix $\boldsymbol \Sigma^i$ (LDA assumes that the covariance matrices are identical across all classes). 

To update our ensemble, we update each base learner $i$ independently using the dataset $X^i$ collected for that learner. To update the parameters of an LDA base learner $i$, let $\boldsymbol z_k$ be a sample from $X^i$ whose label is class $k$. The updated mean vector for class $k$, namely $\widetilde{\boldsymbol \mu}_k^i$, and the updated covariance matrix $\widetilde{\boldsymbol \Sigma}^i$ for the learner are calculated as follows:
\begin{gather}
\widetilde{\boldsymbol \mu}^i_k = \frac{n^i_k \times \boldsymbol \mu^i_k + \boldsymbol z_k}{n^i_k + 1} \\ 
\widetilde{\boldsymbol \Sigma}^i = \frac{N^i}{N^i+1}\boldsymbol \Sigma^i + \frac{1}{N^i+1} \times \frac{n^i_k}{n^i_k + 1}(\boldsymbol z_k-\boldsymbol \mu^i_k)(\boldsymbol z_k-\boldsymbol \mu^i_k)^\top
\end{gather}
where $N^i$ is the number of training samples for the base learner $i$ so far, and $n^i_k$ is the number of training samples for the base learner $i$, labeled as class $k$ so far.

\section{Experiments}\label{sec:semi_supervised_experiments}
We performed experiments with five chronic stroke survivors having hemiparesis and moderate muscle tone: Modified Ashworth Scale (MAS) scores $\leq$ 2 in the upper extremity. Our MAS criteria exclude subjects whose fingers are difficult to move passively --- fingers with more severe spasticity cannot be quickly extended with external forces without increasing muscle tone and risking damage to the joints. Our participants
can fully close their hands but are unable to completely extend their fingers without assistance. The passive range of motion in the fingers is within functional limits. Testing was approved by the Columbia University Institutional Review Board (IRB-AAAS8104) and was performed in a clinical setting under the supervision of an Occupational Therapist. We hypothesize that semi-supervised controls allow us to compensate for intrasession concept drift while maintaining high accuracies.

\subsection{Data Collection} \label{par:data_collection}

We have two data collection protocols: a complete protocol and an abbreviated protocol. For the complete protocol, we collect data for all three intents under different conditions: 1) with the arm resting on a table and the orthosis motor off \{\textit{arm on table motor off}\}, 2) with the arm raised above the table and the orthosis motor off \{\textit{arm off table motor off}\}, and 3) with the arm raised above the table and the orthosis motor on, actively moving the hand \{\textit{arm off table motor on}\}. Specifically for the third case, Table~\ref{tab:modified_train} shows the commands given to the subject and the ground truth intent collected while the motor retracts and extends the tendons.  In the abbreviated protocol, we only collect data from condition three in the complete protocol. For each condition, we ask the subject to open and close their hand three to four times. 
 
\begin{table}
\centering
\caption{Training protocol: for each combination
 of instruction and exotendon state, the table shows the assigned label. We begin with the tendon extended and the subject relaxing (top row, middle column) and proceed 
 counter-clockwise.}
\begin{tabular}{c|ccc}
Device State & \multicolumn{3}{c}{Subject Instruction}\\
&Open&Relax&Close\\\hline\hline
Tendon extended 	& O & R & C \\
Tendon retracting 	& O &   & C \\
Tendon retracted 	& O & R & C 
\end{tabular}

\label{tab:modified_train}
\end{table}

During data collection, the experimenter collects the true subject intent, or ground truth, while providing verbal commands to the subject. For conditions where the motor is on, we move the motor approximately one second after the verbal command is given.

\subsection{Methods} 

In our experiments, we test whether a classifier can be trained using less data and still achieve high accuracies during intent detection. We collect one dataset using the abbreviated protocol. We collect four datasets using the complete protocol --- one trains the \textit{SE-full} baseline control, and the others are testing datasets. Specifically, we evaluate four methods:

\subsubsection{\textbf{Supervised EMG, full training data (SE-full)}} the only classifier trained on labeled data from the complete collection protocol, including data from all three conditions. Classifiers trained under the same conditions as the test data are expected to have high accuracies. We consider this classifier as a baseline, as the assumption of labeled data from the complete protocol is impractical due to high burden on the patient. $\eta=1$ and $\boldsymbol f=[e^1, ... , e^8]^\top$. Classifier parameters do not change after training.

\subsubsection{\textbf{Supervised EMG, partial training data (SE-partial)}} trained on the abbreviated protocol. $\eta=1$, and $\boldsymbol f=[e^1, ... , e^8]^\top$. Classifier parameters do not change.

\subsubsection{\textbf{Supervised Multimodal, partial training data (SM-partial)}} trained on the abbreviated protocol but uses multimodal sensing. $\eta$ is a random number between 5 and 10. The features for each base learner are selected randomly from $\boldsymbol x$. The parameters of this classifier do not change.

\subsubsection{\textbf{Disagreement Semi-Supervised Multimodal, partial training data (DSSM-partial)}} this is our approach, only requiring labels on the abbreviated protocol. Initially, this classifier is the same as \textit{SM-partial}. However, as new data arrives, it is labeled by the disagreement-based oracle and used to update the classifier. 





\begin{table*}
    \renewcommand{\arraystretch}{1.3}
    \caption{Classification accuracy and standard deviation in percentage (\%) for 5 stroke subjects. For each subject, we also provide their genders and ages. We also report the average accuracy across all subjects. The best result is in bold-text. We perform a one-sided Wilcoxon rank sum test on the aggregated results from all subjects and show the computed $p$-values for pairwise differences between \textit{DSSM-partial} and  the  other methods.}
    \label{tab:intrasession_results}
    \centering
    \begin{tabular}{c|ccccc|c|c}
    \toprule
    \textbf{Controls} & 
    \begin{tabular}[c]{@{}c@{}}\textbf{Subject S1}\\Female, 83 \end{tabular} & 
    \begin{tabular}[c]{@{}c@{}}\textbf{Subject S2}\\Male, 71 \end{tabular} 
    & 
    \begin{tabular}[c]{@{}c@{}}\textbf{Subject S3}\\Female, 51 \end{tabular} & 
    \begin{tabular}[c]{@{}c@{}}\textbf{Subject S4}\\Male, 29 \end{tabular} &
    \begin{tabular}[c]{@{}c@{}}\textbf{Subject S5}\\Male, 51 \end{tabular} &
    \textbf{Average} &
    \begin{tabular}[c]{@{}c@{}}\textbf{$p$-value w\slash}\\\textbf{DSSM-partial} \end{tabular}\\
    \midrule
    SE-full & 72.8 $\pm$ 8.1 & 85.6 $\pm$ 4.6 & 63.1 $\pm$ 8.05 & 69.8 $\pm$ 0.81 & 73.8 $\pm$ 3.05 & 73.0 & $3\mathrm{e}{-4}$ \\
    SE-partial & 72.1 $\pm$  4.6 & 81.9 $\pm$ 10.3 & 62.9 $\pm$ 7.92 & 69.8 $\pm$ 7.82 & 70.7 $\pm$ 5.08 & 71.5 & $1\mathrm{e}{-4}$ \\
    SM-partial & 72.7 $\pm$ 5.7 & 80.3 $\pm$ 5.4 & 71.0 $\pm$ 3.76 & 68.3 $\pm$ 0.35 & 73.4 $\pm$ 2.23 & 73.2 & $6\mathrm{e}{-4}$ \\   
    DSSM-partial & \textbf{79.2 $\pm$ 4.4} & \textbf{85.8 $\pm$ 4.0} & \textbf{71.8 $\pm$ 4.10} & \textbf{76.5 $\pm$ 2.65} & \textbf{82.9 $\pm$ 1.34} & \textbf{79.3} & --- \\
    \bottomrule
    \end{tabular}
\end{table*}

We use the above classifiers to predict the subjects' intent for the three testing datasets collected with the complete protocol. We select the multimodal features randomly, but in our experience, this does not notably affect performance. 

\section{Results and Discussion}
For each subject and method, we report the results as mean and standard deviation across the three testing datasets. We convert the ground truth and predicted three-class intent into motor commands as described in Section \ref{sec:intent_detection}. We report the resulting motor-command accuracy: the number of time points at which the command to the motor is correct, divided by the number of total time points. We report motor-command accuracy instead of global accuracy (how often the classifier gets each intent class correct) because we care less about the correct intent than about moving the orthosis as the subject intends.

As shown in Table~\ref{tab:intrasession_results}, our proposed method outperforms all other methods, even over the baseline classifier using complete protocol data. This suggests that semi-supervised learning could make our control more robust to intrasession drift while reducing the training burden. To examine the statistical significance in the difference between our proposed method and the other methods, we perform a one-sided Wilcoxon rank-sum test \cite{wilcoxon1992individual} on the results aggregated across all subjects, using a hypothesis threshold $\alpha = 0.01$. We choose a non-parametric statistical test because we do not assume an underlying normal distribution. Table~\ref{tab:intrasession_results} shows the computed $p$-values for pairwise differences between our proposed method and the other methods. We find all $p$-values to be $< 0.01$; thus, we conclude that the difference between \textit{DSSM-partial} and others indicates a statistically significant improvement in prediction accuracy. 


In the experiments, contrary to our expectations, \textit{SE-full} is not drastically better than \textit{SE-partial}. This is particularly true for subjects S1, S3, and S4. For some stroke subjects, even though we collect datasets following the same protocol, there could be some drift between datasets, caused by subject fatigue, device migration, etc. Given a large drift, the benefit of having more training data is limited. \textit{SE-full} is trained with complete protocol on three conditions (\{\textit{arm on table motor off}\}, \{\textit{arm off table motor off}\}, and \{\textit{arm off table motor on}\}), and the \textit{SE-partial} is trained with only \{\textit{arm off table motor on}\}. It is possible that \textit{SE-partial} is generalizing well enough, even without data from the other two conditions.

The comparison of \textit{SM-partial} and \textit{SE-partial}, which are both trained on a dataset collected with abbreviated protocol, highlights the importance of multimodal sensing and ensemble methods. Ensemble methods have been shown to improve the robustness of machine learning algorithms when large uncertainty is presented and having multiple sensing modalities makes its advantages more pronounced. Even without the disagreement-based updates, the classifier accuracy is improved with multimodal data and an ensemble of base learners. The comparison of \textit{DSSM-partial} and \textit{SM-partial} demonstrates the importance of disagreement-based semi-supervised updates. Despite being trained on partial data, \textit{DSSM-partial} learns from new unlabeled data using semi-supervision and further improves its performance.

\begin{figure*}
\centering
\includegraphics[width=\textwidth]{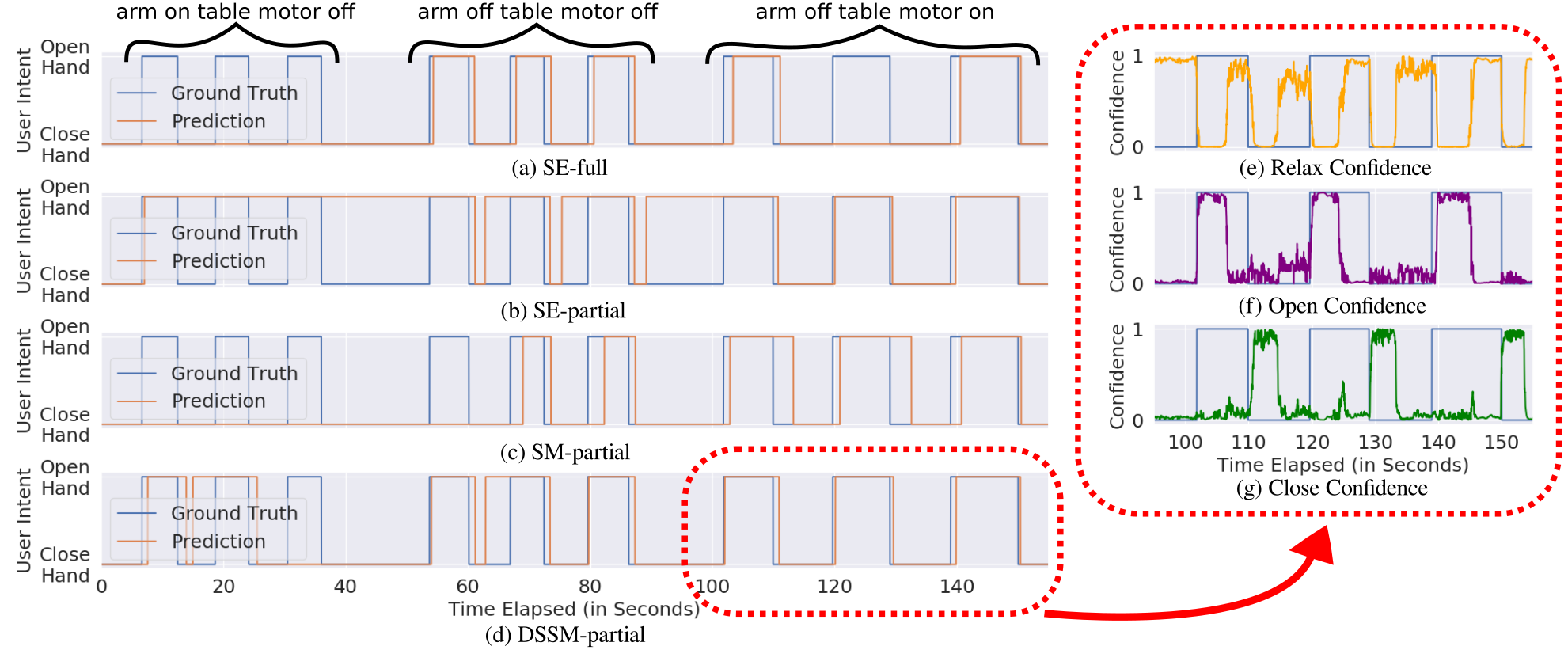}
\caption{Example of the classifiers' output for subject S5 on one of the testing datasets. (a)(b)(c)(d) Comparison of the ground truth user intent and predicted user intent. If the predicted intent is to relax, the intent from the previous time step is used. Data collecting conditions are labeled on top. \textit{DSSM-partial}, despite being trained only on the third condition \{\textit{arm off table motor on}\}, makes correct predictions on the first two conditions and is able to improve its prediction quality as the algorithm runs. (e)(f)(g) Visualization of the confidence produced by \textit{DSSM-partial} on the third condition. The blue line shows the ground truth user intent as in (a)(b)(c)(d), and other colored lines are confidence values.}
\label{fig:prediction}
\end{figure*}

Figure~\ref{fig:prediction} shows a qualitative example of the classifiers' output for subject S5 on one of the testing datasets. Firstly, we notice that prediction on condition \{\textit{arm on table motor off}\} is the most challenging for all methods in this example. This matches our intuition, as this condition is the most functionally different in the protocol. Even \textit{SE-full}, whose training dataset includes this condition, does not successfully predict the intent. However, our proposed method, despite being trained only on the third condition, is able to make correct predictions. Secondly, Figure~\ref{fig:prediction}(d) shows that our proposed method is gradually improving its prediction quality as the algorithm runs. The matching between ground truth and the prediction gets better with semi-supervision, which does not happen in other methods. Figure~\ref{fig:prediction}(e)(f)(g) visualize the confidence of the three-class intent prediction before it is converted to the motor command. The classifier sometimes is confident on a wrong intent, but by setting the threshold for each intent class, we are able to eliminate part of those noisy spikes.

Overall, our experiments show that semi-supervised learning has the potential to reduce the burden placed on users by controls that require an extensive labeled dataset to detect intent. Our proposed algorithm provides an improvement in accuracy when the classifier is trained on a more limited dataset and even outperforms the classifier trained on a larger dataset.

\begin{figure*}
\centering
\includegraphics[width=\textwidth]{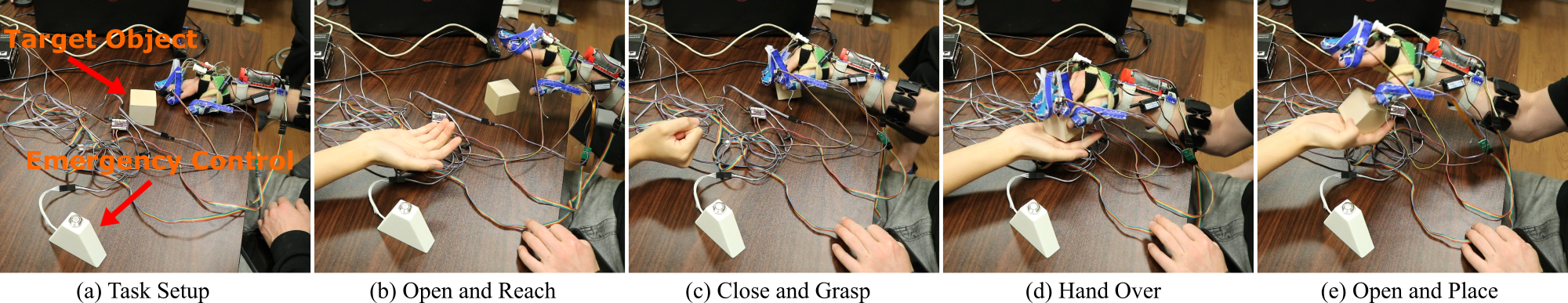}
\caption{Example of subject S5 performing pick-and-handover functional task. (a) The subject is instructed to pick up the wooden block and hand it over to an experimenter. We also provide a button for emergency override of classifier control. The orthosis tendon retracts (hand opens) when the button is pressed and the tendon extends (hand closes) when the button is released. (b) An open signal is detected, the tendon retracts and the subject tries to reach the target object. (c) A close signal is detected, the tendon extends and the subject grasps the block. (d) The subject moves the hand carrying the block to the experimenter. (e) An open signal is detected, the tendon retracts and the subject places the block.}
\label{fig:functional}
\end{figure*}

\section{Functional Tasks}
We conducted a pick-and-handover functional task with subjects S4 and S5 to validate the feasibility of using our proposed algorithm in real time on the hand orthosis. With the orthosis running the disagreement-based semi-supervised algorithm, we instructed each subject to reach and lift a wooden block, then hand over the block to an experimenter who replaced the block back on the table. We continuously repeated this task until one minute elapsed. Both subjects were able to use the orthosis to pick up and hand over the block seven times in a minute. A demonstration of the functional task is shown in Figure~\ref{fig:functional}, and a recording of this experiment can be found in the accompanying video\footnote{The video and more information can be found at \mbox{\url{https://roamlab.github.io/dssm}}}. In the case of subject S4, we observe a number of stops and starts for the first several grasping motions, but this issue is soon alleviated and the control becomes smoother in the later grasping motions, possibly due to adaptive updates. In informal feedback, both subjects found the control to be intuitive and responsive, although we have not quantified this impression using a standardized questionnaire. 

\section{Conclusions and Future Work}

In this paper, we propose a disagreement-based semi-supervised algorithm for addressing intrasession concept drift. Our offline experiment results for five stroke subjects show that semi-supervision helps our controls adapt to intrasession concept drift and could reduce the training burden placed upon the user for orthotic controls. We conduct a functional task with two subjects with the online algorithm, whose success suggests the feasibility of using our method in real time. To our knowledge, this is the first time a semi-supervised learning algorithm has been proposed and used for a hand orthosis based on multimodal ipsilateral sensing. We are also the first to use the proposed algorithm for functional tasks. Our experiment results show that semi-supervised learning is a promising avenue of exploration for orthotic hand controls. They also indicate areas of future research, including handling intersession concept drift that happens between two different sessions, primarily caused by donning and doffing the device.

\newpage
\bibliographystyle{IEEEtran}
\bibliography{bib/related_work,bib/methods}

\end{document}